\documentclass[10pt,twocolumn,letterpaper]{article}

\usepackage[pagenumbers]{cvpr} % To force page numbers, e.g. for an arXiv version

\usepackage[accsupp]{axessibility}
\usepackage{graphicx}
\usepackage{amsmath}
\usepackage{amssymb}
\usepackage{booktabs}

\usepackage{multirow}
\usepackage[table,xcdraw]{xcolor}
\usepackage{amssymb}
\usepackage{bbding}

\usepackage{caption}

\usepackage[ruled,vlined]{algorithm2e} % 算法
\usepackage{float}

\makeatletter
\newcommand{\removelatexerror}{\let\@latex@error\@gobble}
\makeatother

\usepackage[pagebackref,breaklinks,colorlinks]{hyperref}

\usepackage[capitalize]{cleveref}
\crefname{section}{Sec.}{Secs.}
\Crefname{section}{Section}{Sections}
\Crefname{table}{Table}{Tables}
\crefname{table}{Tab.}{Tabs.}

\def\cvprPaperID{4490} % *** Enter the CVPR Paper ID here
\def\confName{CVPR}
\def\confYear{2023}

\begin{document}
\title{Hierarchical Supervision and Shuffle Data Augmentation for \\ 3D Semi-Supervised Object Detection}

\author{\hspace*{-10pt}Chuandong Liu\textsuperscript{\rm 1,2 }, \hspace*{4pt} % \hspace*{-16pt}
Chenqiang Gao\textsuperscript{\rm 1,2}\thanks{Corresponding author.}, \hspace*{4pt} 
Fangcen Liu\textsuperscript{\rm 1,2 },  \hspace*{4pt}
Pengcheng Li\textsuperscript{\rm 1,2 },  \hspace*{4pt}
Deyu Meng\textsuperscript{\rm 3,4 },  \hspace*{4pt}
Xinbo Gao\textsuperscript{\rm 1 }  \\
    $^1$School of Communication and Information Engineering, Chongqing University of Posts and \\
    Telecommunications, Chongqing, China \\
    $^2$Chongqing Key Laboratory of Signal and Information Processing, Chongqing, China\\
    $^3$Xi'an Jiaotong University, Xi'an, China \\
    $^4$Macau University of Science and Technology, Taipa, Macau
}

\maketitle

\begin{abstract}
   State-of-the-art 3D object detectors are usually trained on large-scale datasets with high-quality 3D annotations.
   However, such 3D annotations are often expensive and time-consuming, which may not be practical for real applications.
   A natural remedy is to adopt semi-supervised learning (SSL) by leveraging a limited amount of labeled samples and abundant unlabeled samples.
   Current pseudo-labeling-based SSL object detection methods mainly adopt a teacher-student framework, with a single fixed threshold strategy to generate supervision signals, which inevitably brings confused supervision when guiding the student network training.
   Besides, the data augmentation of the point cloud in the typical teacher-student framework is too weak, and only contains basic down sampling and flip-and-shift (i.e., rotate and scaling), which hinders the effective learning of feature information.
   Hence, we address these issues by introducing a novel approach of Hierarchical Supervision and Shuffle Data Augmentation (HSSDA), which is a simple yet effective teacher-student framework.
   The teacher network generates more reasonable supervision for the student network by designing a dynamic dual-threshold strategy.
   Besides, the shuffle data augmentation strategy is designed to strengthen the feature representation ability of the student network.
   Extensive experiments show that HSSDA consistently outperforms the recent state-of-the-art methods on different datasets. The code will be released at \href{https://github.com/azhuantou/HSSDA}{https://github.com/azhuantou/HSSDA}.
\end{abstract}

%%%%%%%%% BODY TEXT

\begin{figure}[t]
    \centering
    \includegraphics[width=8.0cm]{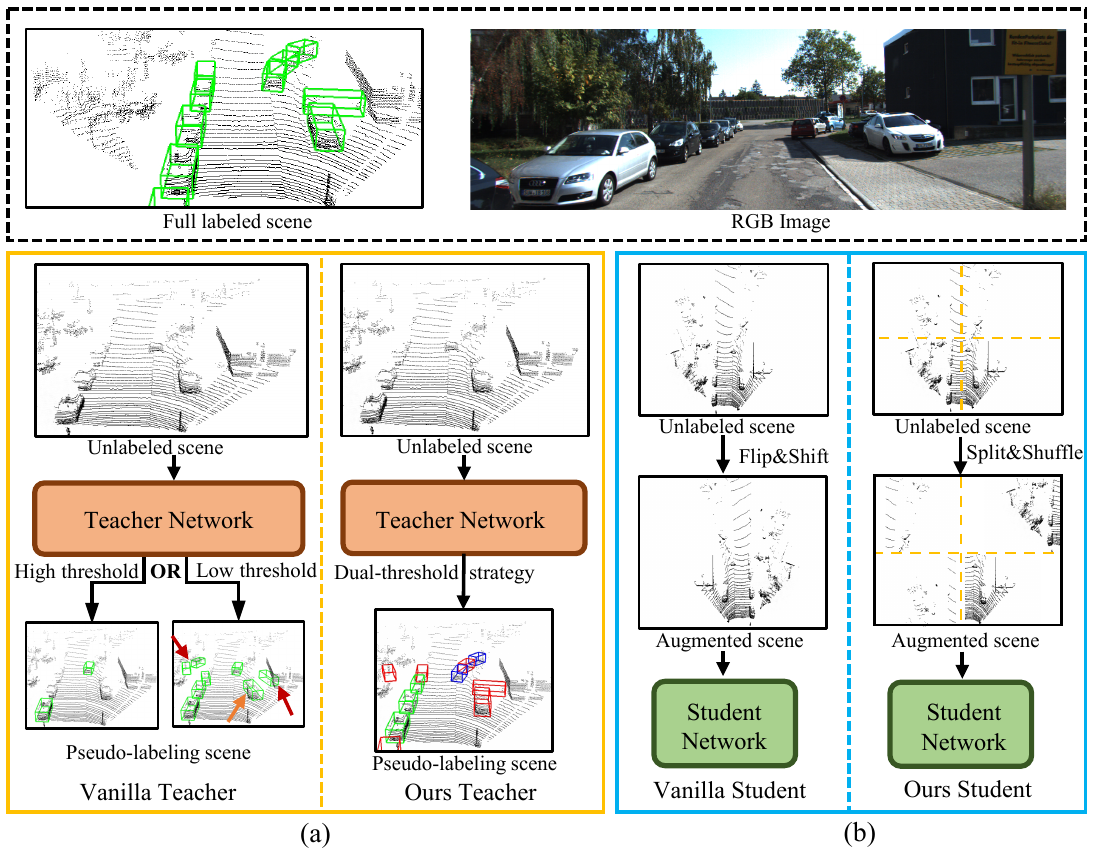}
    \caption{Illustration of (a) the previous teacher compared to our teacher framework and (b) the previous student compared to our student framework. 
    The black dashed box includes the RGB image and the corresponding fully annotated 3D point cloud (green box). 
    The left side of the yellow dotted line in (a) represents the pseudo-labeling scene generated by the single threshold of the vanilla teacher network, causing the student network may be severely misled due to missing mined objects (high threshold) or false positive objects (low threshold), while our proposed teacher network generates three groups of pseudo labels(shown as green, red, blue) to provide hierarchical supervision for the student network. 
    (b) shows our student network adopts stronger shuffled data augmentation than the vanilla student network to learn the stronger ability of feature representation.
    }
    \label{fig:intro}
    \vspace{-1.4em}
\end{figure}

\section{Introduction}
\label{sec:intro}
Kinds of important applications, especially autonomous driving, have been motivating the rapid development of 3D object detection by the range sensor data (\eg, LiDAR point cloud).
Up to now, many point-based and point-voxel-based methods \cite{pointrcnn,pointdensity,pvrcnn++,notallpoints} have been proposed.
Despite the impressive progress, a large amount of accurate instance-level 3D annotations have to be provided for training, which is more time-consuming and expensive than 2D object annotation.
This hinders the application and deployment of existing advanced detection models.
To this end, how to reduce the dependence on huge annotated datasets has achieved growing interest in the object detection community.

As one of the important machine learning schemes for reducing data annotation, semi-supervised learning (SSL) aims to improve the generalization ability of model training with a small number of labeled data together with large-scale unlabeled samples.
In the semi-supervised object detection community, most works focus on 2D object detection \cite{stac,jeong2019consistency-2d-semi,unbiasedv2,wang2021data-2d-semi,softteacher}, among which the teacher-student model is the mainstream framework. 
Specifically, the teacher network with weakly augmented labeled data generates pseudo labels to train the student network with strong data augmentation. 
This pipeline has been widely verified to be effective, in which the quality of pseudo labels and the data augmentation strategy are the key and many works have been proposed to tackle them \cite{fixmatch,remixmatch,cutout,stac}. 
Benefiting from 2D semi-supervised object detection, several 3D semi-supervised object detection methods have been proposed \cite{3dioumatch,sess,detmatch,proficient_teachers}, which still mainly adopted the teacher-student model. 
These methods, as well as 2D semi-supervised object detection methods \cite{stac,instantteacher,activeteacher}, mainly use a hard way, \eg, a score threshold, to get pseudo labels to train the student network. 
This kind of strategy is difficult to guarantee the quality of pseudo labels. Taking the score threshold strategy as an example, if the threshold is too low, the pseudo labels will contain many false objects, while if it is too high, the pseudo labels will miss many real objects which will be improperly used as background (see in \cref{fig:intro} (a)). 
As exhibited in \cite{3dioumatch}, only about 30\% of the objects can be mined from the unlabeled scenes even at the end of the network training. 
Thus, both of those  two cases will bring the student network confused supervision, which harms the performance of the teacher-student model. This would inevitably happen for the single threshold strategy, even adopting some optimal threshold search method \cite{3dioumatch}. 
Thus, how to produce reasonable pseudo labels from the teacher network output is an important issue to address for better training the student networks. 

Besides the quality of pseudo labels, data augmentation is also the key to the teacher-student model as mentioned previously. 
Extensive works in 2D semi-supervised object detection have shown that strong data augmentation is very important to learn the strong feature representation ability of the student network. 
Thus, kinds of strong data augmentation strategies, \eg, Mixup \cite{mixup}, Cutout \cite{cutout}, and Mosaic~\cite{mosaic} have been widely adopted. 
However, current 3D semi-supervised object detection methods adopt some weak data augmentation strategies, \eg, flip-and-shift. 
These kinds of data augmentations are not able to well drive the student network to learn strong feature representation ability. 
Thus, the good effect of data augmentation in 2D semi-supervised object detection does not appear obviously in 3D semi-supervised object detection. 

To tackle the above issues of the quality of pseudo labels and data augmentation, we propose a Hierarchical Supervision and Shuffle Data Augmentation (HSSDA) method for 3D semi-supervised object detection.
We still adopt the teacher-student model as our mainframe. 
For obtaining more reasonable pseudo labels for the student network, we design a dynamic dual-threshold strategy to divide the output of the teacher network into three groups: (1) high-confidence level pseudo labels, (2) ambiguous level pseudo labels, and (3) low-confidence level pseudo labels, as shown in \cref{fig:intro} (a). 
This division provides hierarchical supervision signals for the student network. 
Specifically, the first group is used as the strong labels to learn the student network, while the second join learning through a soft-weight way. 
The higher the score is, the more it affects learning. 
The third group is much more likely to tend to be false objects. 
We directly delete them from the point cloud to avoid confusing parts of the object point cloud into the background. 

For strengthening the feature representation ability of the student network, we design a shuffle data augmentation strategy in this paper. 
As shown in \cref{fig:intro} (b), we first generate shuffled scenes by splitting and shuffling the point cloud patches in BEV (bird-eye view) and use them as inputs to the student model.
Next, the feature maps extracted from the detector backbone are unshuffled back to the original point cloud geometry location.

To summarize, our contributions are as follows:
\begin{itemize}
\item We propose a novel hierarchical supervision generation and learning strategy for the teacher-student model. This strategy can provide the student network hierarchical supervision signal, which can fully utilize the output of the teacher network.
\item We propose a shuffle data augmentation strategy that can strengthen the feature representation ability of the student network. 
\item Our proposed hierarchical supervision strategy and shuffle data augmentation strategy can be directly applied to the off-the-shelf 3D semi-supervised point cloud object detector and extensive experiments demonstrate that our method has achieved state-of-the-art results.
\end{itemize}

% -------------------------Related work-----------------------------------

\section{Related Work}
\label{sec:Related Work}

\subsection{3D Object Detection}
3D object detection is a fundamental task in the autonomous driving area.
The mainstream 3D object detection methods can be roughly divided into three types: voxel-based methods \cite{second,sessd,ciassd,pointpillars,votr,voxset}, point-based methods \cite{pointrcnn,std,3dssd,vote,pointgnn,pcrgnn}, and point-voxel-based methods\cite{sassd,pvrcnn,pvrcnn++,ct3d,pyramidrcnn}. 
For voxel-based methods, voxelization is a common operation that transforms irregular point clouds into voxel grids for applying traditional 2D or 3D convolution. 
In Voxelnet \cite{voxelnet}, the voxel-wise encoding layer was adopted for collective feature representation extraction from the voxel-wise LiDAR point cloud.
VoxSeT \cite{voxset} presented a novel transformer-based framework that encoded features from a larger receptive field.
Point-based approaches directly use the raw point cloud to capture spatial structure information for feature extraction through networks of the PointNet series \cite{pointnet, pointnet++}.
PointRCNN \cite{pointrcnn} is the representative, which directly generates point-level RoIs and uses the point-level features for further refinement.
To accelerate the inference speed for applications, IA-SSD \cite{iassd} proposed an efficient downsampling way and a contextual centroid perception module to capture geometrical structure.
Point-Voxel-based methods combined voxel representations with point representations from the point cloud.
Built on the PV-RCNN \cite{pvrcnn}, PV-RCNN++ \cite{pvrcnn++} leveraged a VectorPool aggregation to learn structure features and a sectorized proposal-centric keypoints sampling strategy to obtain more keypoints.
All the above fully supervised methods can be easily embedded into our HSSDA framework, \eg, PV-RCNN \cite{pvrcnn} and Voxel-RCNN \cite{voxelrcnn}.

\subsection{Semi-supervised Learning (SSL)}
SSL can greatly reduce the annotations for model training and most existing works focus on image classification, which can be broadly divided into two types: consistency regularization \cite{mixmatch,remixmatch,consistency_1,consistency_2} and pseudo-labeling methods \cite{pseudo_label_1,pseudo_label_2,selftraining}.
The former approaches assume the model's predictions to be consistent under input perturbations/augmentations (\eg, different contrast, flip, etc.) and penalize the inconsistency of predictions.
Techniques range from simple augmentation to more complex transformations such as MixUp \cite{mixup}, as well as stronger automatic augmentation such as Cutout \cite{cutout} and CTAugment \cite{remixmatch}.
The latter methods exploit pseudo-labeling, where the model first is trained trains on labeled data and then iteratively generates the pseudo labels of unlabeled data to add highly confident predictions for training.
It is revisited in deep neural networks to learn from large amounts of unlabeled data.
Notably, perturbation mechanisms of the above two types of methods play a key role in promoting the model robustness against noise in network parameters or structure but have not been explored in SSL for 3D object detection.

\subsection{Semi-supervised Object Detection}
Inspired by the SSL works in image classification, SSL is also applied to the 2D object detection to alleviate the heavy annotation problem \cite{stac,jeong2019consistency-2d-semi,unbiasedv2,wang2021data-2d-semi}.
STAC \cite{stac} generated pseudo labels for unlabeled data in an offline manner. 
To further improve the quality of pseudo labels, Instant-Teachering \cite{instantteacher} rectified the false predictions via the co-rectify scheme and experimented with MixUp \cite{mixup} and Mosaic \cite{mosaic}. 
This work aims to tackle a more challenging task, SSL for 3D object detection, where large spaces of 7 Degrees-of-Freedom of 3D objects need to be searched.

Recently, several works have been proposed in the 3D SSL domain.
SESS \cite{sess} and 3DIoUMatch \cite{3dioumatch} are the pioneer approaches for 3D object detection from indoor and outdoor point cloud data.
Similar to 2D SSL methods, SESS \cite{sess} leveraged a triple consistency regularization strategy to align the 3D proposal from the teacher and student network.
Following the pseudo-labeling line in SSL, 3DIoUMatch \cite{3dioumatch} designed a series of filtering strategies such as objectness, classification, and localization threshold to obtain high-quality pseudo labels, and a unique IoU estimation branch to further deduplicate the predictions.
Different from 3DIoUMatch, Proficient Teachers \cite{proficient_teachers} developed several necessary modules to improve the recall and precision of pseudo labels and removed the necessity of threshold setting.
DetMatch \cite{detmatch} generated more precise pseudo labels by matching 2D and 3D detection from each modality.
These works employed the EMA weight update strategy to train a student network and then gradually update the teacher network. %  in a mutually-beneficial manner.
Although achieving impressive performances with high-quality pseudo labels, the missing-mined objects and heuristic strong augmentation are ignored.
By contrast, our HSSDA leverages the hierarchical supervision and shuffle data augmentation to alleviate these issues and further improve performance.

% -----------------methods--------------------------------------------------

\begin{figure*}[t]
    \centering
    \includegraphics[width=15.3cm]{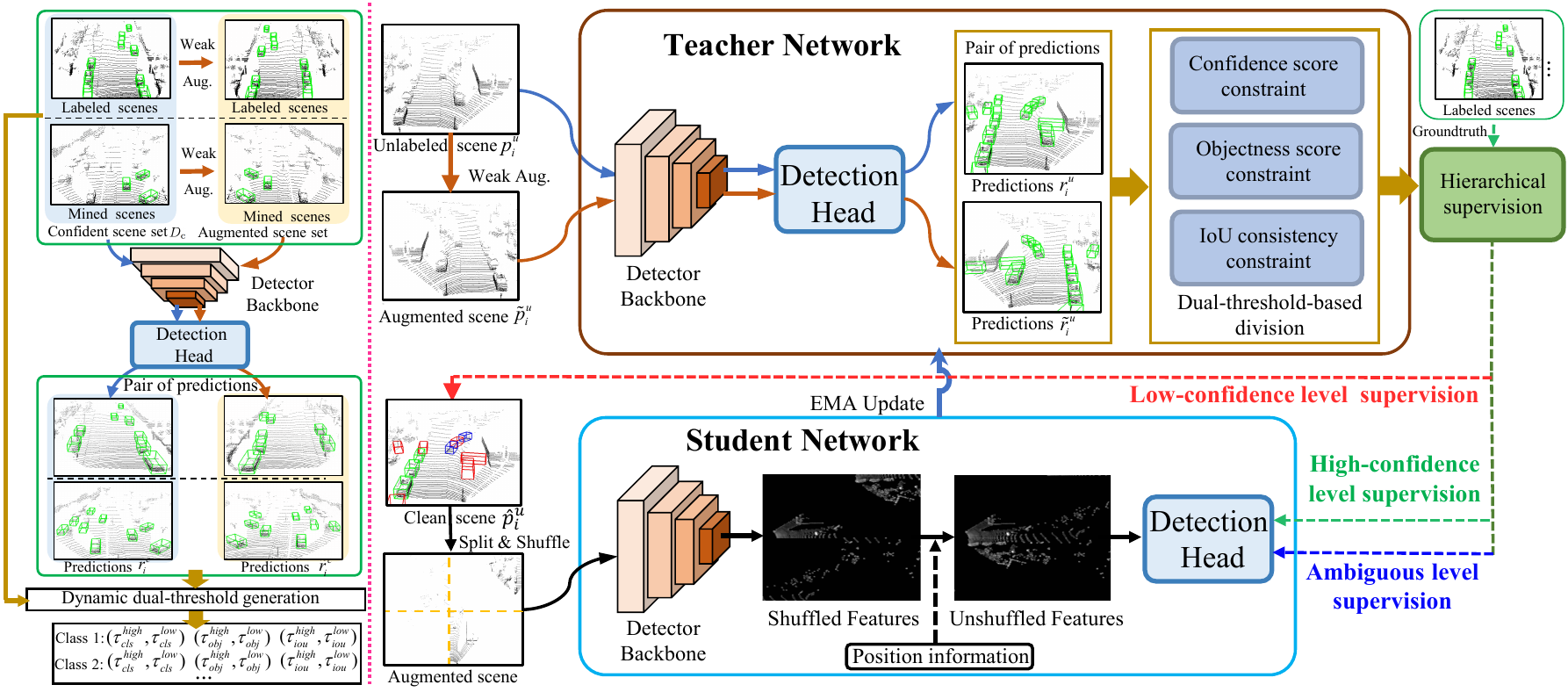}  % 16.3
    \caption{
    Overview of the proposed HSSDA pipeline.
    % Our approach is implemented based on the teacher-student framework.
    We propose a dual-threshold strategy to help the teacher network to generate hierarchical supervision to train the student network. 
    Besides, we also propose a data augmentation strategy to strengthen the ability of feature representation of the student network.
    }
    \label{fig:pipline}
    \vspace{-1.0em}
\end{figure*}

\section{Method}

\subsection{Preliminary}
\label{sec:3.2}
\textbf{Problem Definition}
We first provide the definition of semi-supervised 3D object detection.
In detail, the model is trained with a set of labeled scenes $D_{\mathrm{s}}=\left\{p_{i}^{\mathrm{s}}, y_{i}^{\mathrm{s}}\right\}_{i=1}^{N_{\mathrm{s}}}$ and a set of unlabeled scenes $D_{\mathrm{u}}=\left\{p_{i}^{\mathrm{u}}\right\}_{i=1}^{N_{\mathrm{u}}}$, where $p_{i} \in \mathcal{R}^{n \times\{3+r\}}$ represents a point cloud scene $p_{i}$ which has $n$ points with three-dimensional coordinates and additional r-dimensional information (\eg, color, intensity) that can be treated as extra features, $N_{\mathrm{s}}$ and $N_{\mathrm{u}}$ are the numbers of labeled and unlabeled point cloud scenes, respectively.
Generally speaking, $N_{u}>>N_{s}$.
For a scene $p_{i}^{\mathrm{s}}$, the annotation $y_{i}^{\mathrm{s}}$ is composed of both 7-dimensional location information which includes center, size, and orientation, and category of the 3D bounding boxes.

\textbf{Teacher-Student Framework}
Similar to the mainstream researches \cite{3dioumatch,sess,pseudoaugment}, our learning paradigm also builds up on the teacher-student framework which includes two 3D detectors with the same configurations.
Here, we can use any off-the-shelf state-of-the-art 3D object detector, \eg, PV-RCNN \cite{pvrcnn} and Voxel-RCNN \cite{voxelrcnn}.
Following those works, we build the teacher detector via exponential moving average (EMA) \cite{meanteacher}:
\begin{align} \label{ema}
  \theta_t^{i+1}=\theta_t^{i} \cdot \alpha+\theta_s^{i} \cdot(1-\alpha), 
\end{align}
where $\alpha$ is the EMA decay rate, $\theta_t$ and $\theta_s$ represent the parameters of the teacher and student networks, respectively, and $i$ denotes the training step.

\subsection{Overview}
\label{sec:overview}
The pipeline of our HSSDA framework is illustrated in \cref{fig:pipline}, which is derived from the basic teacher-student mutual learning framework.
In the burn-in stage of training, we train the detector in a fully supervised manner following OpenPCDet \cite{openpc} with the labeled scenes and keep the same setting as the used detector.
Then, both the teacher network and student network are initialized with the same pre-trained weight parameters.

In the mutual learning stage, there are three steps in each training epoch.
The first step is to generate three kinds of dual-thresholds for each class in a global view, as shown on the left of the pink dotted line in \cref{fig:pipline}.
Specifically, we construct a confident scene set $D_{\mathrm{c}}$ composed of labeled scenes and mined scenes (in the first epoch, it just contains all labeled scenes).
% , whilst its weak augmentation (rotation and scaling) is sent to the teacher network for predictions.
Then we sequentially input each scene from $D_{\mathrm{c}}$ and its weak augmentation (rotation and scaling) into the teacher network to produce a pair of predictions. 
Based on those pairs of predictions and the object information from $D_c$, we design a dynamic dual-threshold generation strategy to obtain three kinds of dual-thresholds for each class in terms of confidence score, objectness score, and IoU consistency: $(\tau_{cls}^{high}, \tau_{cls}^{low})$, $(\tau_{obj}^{high}, \tau_{obj}^{low}$), $(\tau_{iou}^{high}$, $\tau_{iou}^{low})$.
The second step (see the right of the pink dotted line in \cref{fig:pipline}) is mainly to mine the hierarchical pseudo labels. 
Specifically, each unlabeled scene $p_i^u$ and its weak augmented scene $\tilde{p}_i^u$ are sequentially input to the teacher detector to produce a pair of predictions.
Through three measure rules based on the dual-thresholds obtained in the first step, we can generate the hierarchical supervision: (1) high-confidence level pseudo labels, (2) ambiguous level pseudo labels, and (3) low-confidence level pseudo labels.
We add all high-confidence pseudo labels into the confident scene set $D_c$ for the next dual-threshold generation.
We also add the ground-truth from labeled scenes into the first group for following student network training. 
In the third step, we use the hierarchical supervision composed of three groups of pseudo labels to train the student network with our designed shuffle data augmentation, and then update the teacher network by the EMA strategy according to \cref{ema}.

After the mutual training step, we use the 3D detector from the student network as our final detector. 
Through the above procedure, we can see that our designed framework can train any off-the-shelf 3D detector which consists of a backbone and a detection head. 

\subsection{Dynamic dual-threshold generation}

\begin{algorithm}[t]
  \caption{Dynamic dual-threshold generation}%算法名字
  \label{alg:dual_threshold}
  \LinesNumbered %要求显示行号
  \KwIn{confident scene set $D_{\mathrm{c}}$, pairs of predictions $r^c$ and $\tilde{r}^c$, IoU matching threshold $\tau_{pair}$.}%输入参数
  \KwOut{$(\tau_{cls}^{high}, \tau_{cls}^{low})$, $(\tau_{obj}^{high}, \tau_{obj}^{low}$), $(\tau_{iou}^{high}$, $\tau_{iou}^{low})$}%输出
%   initialize confidence set $D_{\mathrm{c}} = D_{\mathrm{s}}$\;
%   \For{$e = 1,2, \ldots,E$}{
  initialize empty sets $\mathcal{P}_{cls}$, $\mathcal{P}_{obj}$, and $\mathcal{P}_{iou}$\;
  \For{each $\left\{p_{i}^{\mathrm{c}}, y_{i}^{\mathrm{c}}\right\} \in D_{\mathrm{c}}$}{
    fetch bounding boxes $b_i^{gt}$ from $y_i^c$\;
    fetch bounding boxes $b_i^{r}$ from $r_i^c$\;
    fetch bounding boxes $\tilde{b}_i^{r}$ from $\tilde{r}_i^c$\;
    \For{$b_{ij}^{gt}$ in $b_i^{gt}$}{
     compute the matrix $\mathcal{M} \leftarrow IoU(b_{ij}^{gt}, b_i^r)$\;
     \If{$max(\mathcal{M}) > \tau_{pair}$}{
      choose index $k$ with $max(\mathcal{M})$\;
      $\mathcal{P}_{cls}=\mathcal{P}_{cls} \cup s^c_{cls}$, \\ 
      $s^c_{cls}$ is the confidence score of $r_{ik}^c$\;
      $\mathcal{P}_{obj}=\mathcal{P}_{obj} \cup s^c_{obj}$, \\ 
      $s^c_{obj}$ is the objectness score of $r_{ik}^c$\;
      $\mathcal{P}_{iou}=\mathcal{P}_{iou} \cup v^c$, \\ 
      $v^c = max(IoU(b_{ik}^r, \tilde{b}_{i}^r))$\;
     }
    }
  }
  $(\tau_{cls}^{high}, \tau_{cls}^{low}), (\tau_{obj}^{high}, \tau_{obj}^{low}), (\tau_{iou}^{high} \tau_{iou}^{low}) \leftarrow JNB(\mathcal{P}_{cls}, \mathcal{P}_{obj}, \mathcal{P}_{iou})$\;
  % \vspace{-1.4em}
\end{algorithm}

\begin{figure}[t]
    \centering
    \includegraphics[width=7.2cm]{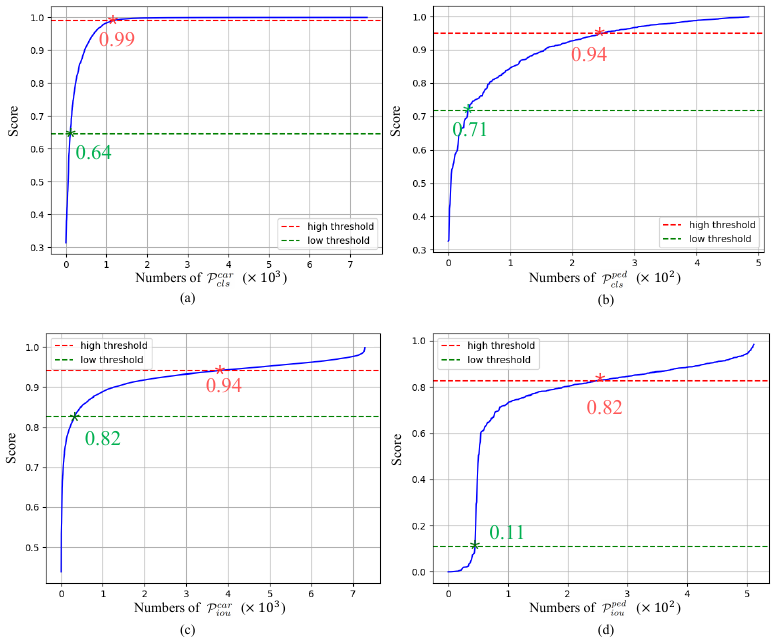}
    \caption{Selection of the dual-threshold based on the Jenks Natural Breaks \cite{jenks}. 
    In each subplot, the blue line represents the sorted scores from the sequence pool $\mathcal{P}$ and the red and green dash lines indicate the high threshold and low threshold, respectively.
    (a) and (b) show the process of dynamic confidence IoU selection for `Car' and `Pedestrian'.
    Similarly, (c) and (d) show the selection of dynamic consistency IoU threshold. 
    }
    \label{fig:threshold}
    \vspace{-1.4em}
\end{figure}

The dual threshold generation is dynamically conducted in each training epoch. 
\cref{alg:dual_threshold} describes the whole process of generating dual-thresholds for one class in one training epoch, given a confidence set $D_{\mathrm{c}} = \left\{p_{i}^{\mathrm{c}}, y_{i}^{\mathrm{c}}\right\}_{i=1}^{N_{\mathrm{c}}}$, its pairs of predictions $r^c$, $\tilde{r}^c$ through the 3D detector, and an IoU matching threshold $\tau_{pair}$. 
Here $\tau_{pair}$ is determined experientially and is used to judge if two 3D bounding boxes match. 
Initially, we create three empty sets $\mathcal{P}_{cls}$, $\mathcal{P}_{obj}$ and $\mathcal{P}_{iou}$ to collect the confidence score, objectness score and IoU. 
These three sets will be used to search three optimal dual-thresholds. 
Concretely, for the $i$-th scene, we can fetch the predicted bounding boxes ${b_{i}^r}$ and ground truth bounding boxes ${b_{i}^{gt}}$ from $r_i^c$ and $y_{i}^c$, respectively.
Further, we utilize the common IoU-based strategy to pair a predicted box ${b_{ik}^r}$ for each ground truth ${b_{ij}^{gt}}$ in the $i$-th scene, aiming to obtain the predicted confidence scores and predicted objectness scores of ground truth objects, and collect these scores into the confidence score set $\mathcal{P}_{cls}$ and objectness score set $\mathcal{P}_{obj}$, respectively.
In this way, we can distinguish predictions with different classifications and objectness reliability in a global view.
At the same time, we can also get the consistency IoU set $\mathcal{P}_{iou}$ based on the $r_i^c$ and $\tilde{r}_i^c$, which facilitates grading predictions with different localization quality in a consistency constraints manner.
After handling all scenes, we solve the dual-threshold search problem through a global clustering algorithm.
Specifically, we adopt the Jenks Natural Breaks (JNB) \cite{jenks} algorithm to search the natural turning points or breakpoints based on the three sets.
As shown in \cref{fig:threshold} (a) and (b), we can automatically obtain the dual-threshold $\tau_{cls}^{high}$ and $\tau_{cls}^{low}$ for each class.
% Moreover, we will update the confidence set $D_{\mathrm{c}}$ by exploiting the unlabeled scenes $D_{\mathrm{u}}$ via \cref{dc}.
Similarly, we can automatically obtain the other two dual-thresholds $\tau_{obj}$ and $\tau_{iou}$, which are detector-agnostic and category-aware.
As mentioned in \cref{sec:overview}, in each training epoch, the confident scene $D_c$ will be updated, so all three dual-thresholds will dynamically change during training.

\subsection{Hierarchical supervision generation}
\label{sec:hs}

As shown in \cref{fig:pipline}, with the three dual-thresholds for each class, we divide the mined object pseudo labels from unlabeled scenes into hierarchical supervision which includes (1) high-confidence level pseudo labels, (2) ambiguous level pseudo labels and (3) low-confidence level pseudo labels. 
Specifically, the high-confidence level pseudo label $\tilde{y}_{i}$ is chosen when the predicted result $r_i^u$ of unlabeled scenes from the teacher network meets all three following inequalities simultaneously, \emph{i.e.,} $s_{cls}^u > \tau_{cls}^{high}$, $s_{obj}^u > \tau_{obj}^{high}$ and $v^u > \tau_{iou}^{high}$, where $s_{cls}^u$ and $s_{obj}^u$ are the confidence score and the objectness score of the chosen predicted result, respectively, and $v^u$ is the consistency IoU between chosen $r_i^u$ and $\tilde{r}_i^u$.
Besides, we group the rest of the predictions as ambiguous level pseudo labels, which meet all three following inequalities simultaneously, \emph{i.e.,} $s_{cls}^u > \tau_{cls}^{low}$, $s_{obj}^u > \tau_{obj}^{low}$ and $v^u > \tau_{iou}^{low}$.
As for those predictions that neither belong to the high-confidence level nor ambiguous level, we mark them as low-confidence level predictions.

As mentioned previously, we also add the ground-truth from labeled scenes into the group of high-confidence level pseudo labels which provide strong object label supervision for the student network, while the ambiguous level group will supervise the student network training through a soft-weight way which will be introduced in \cref{sec:training}. 
Following~\cite{ss3d}, we leverage the points removal strategy to eliminate noise information based on the group of low-confidence level pseudo labels.

\subsection{Shuffle Data Augmentation} 

Weak-strong data augmentation plays a significant role in the teacher-student style framework, which guides the model to learn strong feature representation. As mentioned previously, the data augmentation of the student network of current methods is too weak to learn the strong ability of feature representation.
However, due to the huge modality difference between 2D images and 3D point cloud, it is not feasible to directly apply the widely-used data augmentation strategies, \eg, color transformation or Mixup \cite{mixup} to point cloud for object detection.
Thus, we propose a shuffle data augmentation strategy for the student network.
Specifically, as shown in \cref{fig:pipline}, given a scene $\hat{p}_i^u$ from the teacher network, we first clip the range of point cloud scene into $[x_1, x_2]$ for the X axis, $[y_1, y_2]$ for the Y axis and compress the Z axis to form BEV (bird-eye-view) grid.
Then we split each scene into $\mathcal{R} \times \mathcal{C}$ (\eg, $2 \times 2$) patches and generate the same shaped $\mathcal{R} \times \mathcal{C}$ position information to shuffle each scene patch.
By feeding the shuffled patches to the detector backbone for feature extraction, the teacher network with a weak branch and the student network with a strong branch can achieve obvious differences, which is beneficial for the student network to learn more complex and peculiar information from hierarchical supervision.
Then we leverage the position information to unshuffle the patch features for further regression and classification in the detection head.
Although shuffled patches make it difficult to distinguish the edges or parts of objects in scenes, the unshuffle operation before the detection head restores the original location from the feature space.
Hence, our student network delivers more effort into learning with weaker features, which would strengthen the ability of feature representation, and especially benefit to detect objects with weak feature due to small sizes, \eg, `Pedestrian', `Cyclist', which will be shown in experiments.

\subsection{Training Objective Function}
\label{sec:training}
Following \cite{humbleteacher,unbiasedv2}, we freeze the optimization of the teacher detector, and the student detector is trained on both unlabeled scenes with the hierarchical supervision and labeled scenes with the ground-truth.
More specifically, the training objective consists of a supervised loss for labeled and unlabeled scenes.
\begin{align} 
  \mathcal{L}_{s}=\sum_{i} \mathcal{L}_{c l s}\left(p_{i}^{s}, y_{i}^{s}\right)+\mathcal{L}_{r e g}\left(p_{i}^{s}, y_{i}^{s}\right), \label{ls} \\
  \mathcal{L}_{u}=\sum_{i} \mathcal{L}_{c l s}\left(\hat{p}_{i}^{u}, \tilde{y}_{ij}^{u}\right)+\mathcal{L}_{r e g}\left(\hat{p}_{i}^{u},\tilde{y}_{ij}^{u}\right) \notag \\ +w_{ij}\mathcal{L}_{c l s}\left(\hat{p}_{i}^{u}, \hat{y}_{ij}^{u}\right)+w_{ij}\mathcal{L}_{r e g}\left(\hat{p}_{i}^{u}, \hat{y}_{ij}^{u}\right), \label{lu}
%   w_{ij} = s_{cls} \cdot s_{obj}, 
\end{align}
where $\mathcal{L}_{cls}$ is the classification loss, $\mathcal{L}_{reg}$ is the regression loss, $\hat{y}_{ij}^{u}$ and $\tilde{y}_{ij}^{u}$ are the $j$-th ambiguous level pseudo labels and high-confidence level pseudo labels generated by the teacher detector in the $i$-th scene, and $w_{ij}=r^{cls}_{ij} \cdot r^{obj}_{ij}$ denotes the soft-weight for ambiguous level pseudo labels $\hat{y}_{ij}^{u}$, which is determined by a combination of predicted confidence score and predicted objectness reliability score.
Thanks to the clean scenes generated by the noise points removal operation and further obtaining complex scenes by GT sampling data augmentation \cite{second}, we do not force each training batch to contain a mixture of labeled scenes $p_i^s$ and unlabeled scenes $p_i^u$ with a certain ratio (\eg, $1:1$ in \cite{3dioumatch}), but randomly sample each batch from the entire dataset for training.
Hence, the training loss is defined as follows:
\begin{align} \label{overall}
    \mathcal{L}=\mathcal{L}_{s}+\mathcal{L}_{u}. 
\end{align}
Thus, we remove the hyper-parameter to trade-off between $\mathcal{L}_{s}$ and $\mathcal{L}_{u}$ as used in the common teacher-student framework\cite{3dioumatch,unbiasedv2,stac}.

\begin{table*}
\centering
\resizebox{\textwidth}{!}{
% \resizebox{13cm}{!}{
\setlength{\abovecaptionskip}{0.cm}
\begin{tabular}{c|c|cccc|cccc|cccc}
\hline
\multirow{2}{*}{Model} & \multirow{2}{*}{Modality} & \multicolumn{4}{c|}{1\%} & \multicolumn{4}{c|}{2\%} & \multicolumn{4}{c}{20\%} \\
 &  & Car & Ped. & Cyc. & Avg. & Car & Ped. & Cyc. & Avg. & Car & Ped. & Cyc. & Avg. \\ \hline \hline
PV-RCNN \cite{pvrcnn} & LiDAR & 73.5 & 28.7 & 28.4 & 43.5 & 76.6 & 40.8 & 45.5 & 54.3 & 77.9 & 47.1 & 58.9 & 61.3 \\ \hline \hline
3DIoUMatch \cite{3dioumatch} (PVR.-based) & LiDAR & 76.0 & 31.7 & 36.4 & 48.0 & 78.7 & 48.2 & 56.2 & 61.0 & - & - & - & - \\ \hline
DetMatch \cite{detmatch} (PVR.\&FR.\cite{fasterrcnn}-based) & LiDAR + RGB & 77.5 & \textbf{57.3} & 42.3 & 59.0 & 78.2 & 54.1 & 64.7 & 65.6 & 78.7 & 57.6 & 69.6 & 68.7 \\ \hline
Our HSSDA (PVR.-based) & LiDAR & \textbf{80.9} & 51.9 & \textbf{45.7} & \textbf{59.5} & \textbf{81.9} & \textbf{58.2} & \textbf{65.8} & \textbf{68.6} & \textbf{82.5} & \textbf{59.1} & \textbf{73.2} & \textbf{71.6} \\ \hline
\end{tabular}}
\setlength{\abovecaptionskip}{0.13cm}
\caption{Experimental results on KITTI dataset compared with recent state-of-the-art methods.
For fair comparison, the results are reported with 40 recall positions, under IoU thresholds 0.7, 0.5, and 0.5 for `Car', `Pedestrian', and `Cyclist', respectively.
}
\label{tab:kitti_result}
\vspace{-0.5em}
\end{table*}

\begin{table*}
\centering
\resizebox{\textwidth}{!}{
% \resizebox{13cm}{!}{
\begin{tabular}{c|c|cc|cc|cc|cc|cc|cc}
\hline
\multirow{2}{*}{\begin{tabular}[c]{@{}c@{}}1\% Data \\ ($\sim$ 1.4k scenes)\end{tabular}} & \multirow{2}{*}{Modality} & \multicolumn{2}{c|}{Veh. (LEVEL 1)} & \multicolumn{2}{c|}{Veh. (LEVEL 2)} & \multicolumn{2}{c|}{Ped. (LEVEL 1)} & \multicolumn{2}{c|}{Ped. (LEVEL 2)} & \multicolumn{2}{c|}{Cyc. (LEVEL 1)} & \multicolumn{2}{c}{Cyc. (LEVEL 2)} \\
 &  & mAP & mAPH & mAP & mAPH & mAP & mAPH & mAP & mAPH & mAP & mAPH & mAP & mAPH \\ \hline \hline
PV-RCNN \cite{pvrcnn} & LiDAR & 47.3 & 45.6 & 43.6 & 42.0 & 28.9 & 15.6 & 26.2 & 14.1 & - & - & - & - \\ \hline
DetMatch \cite{detmatch} (PVR.\& FR.\cite{fasterrcnn}-based) & LiDAR+RGB & 52.2 & 51.1 & 48.1 & 47.2 & 39.5 & 18.9 & \textbf{35.8} & 17.1 & - & - & - & - \\
Improvement & - & +4.9 & +5.5 & +4.5 & +5.2 & +10.6 & +3.3 & +9.6 & +3.0 & - & - & - & - \\ \hline \hline
PV-RCNN \cite{pvrcnn} (Reproduced by us) & LiDAR & 48.5 & 46.2 & 45.5 & 43.3 & 30.1 & 15.7 & 27.3 & 15.9 & 4.5 & 3.0 & 4.3 & 2.9 \\ \hline
Our HSSDA (PVR.-based) & LiDAR & \textbf{56.4} & \textbf{53.8} & \textbf{49.7} & \textbf{47.3} & \textbf{40.1} & \textbf{20.9} & 33.5 & \textbf{17.5} & \textbf{29.1} & \textbf{20.9} & \textbf{27.9} & \textbf{20.0} \\
Improvement & - & +7.9 & +7.6 & +4.2 & +4.0 & +10.0 & +5.2 & +6.2 & +1.6 & +24.6 & +17.9 & +23.6 & +17.1 \\ \hline
\end{tabular}}
\setlength{\abovecaptionskip}{0.13cm}
\caption{Performance comparison on the Waymo Open Dataset with 202 validation sequences for the 3D detection.
}
\label{tab:waymo_result}
\vspace{-1.0em}
\end{table*}

\begin{table}
\centering
% \resizebox{\textwidth}{!}{
% \resizebox{7cm}{!}{
\resizebox{\columnwidth}{!}{
\begin{tabular}{c|c|ccc|ccc|ccc}
\hline
 &  & \multicolumn{3}{c|}{3D Detection (Car)} & \multicolumn{3}{c|}{3D Detection (Ped)} & \multicolumn{3}{c}{3D Detection (Cyc)} \\ \cline{3-11} 
\multirow{-2}{*}{Model} & \multirow{-2}{*}{Data} & Easy & Mod & Hard & Easy & Mod & Hard & Easy & Mod & Hard \\ \hline \hline
Voxel-RCNN \cite{voxelrcnn} & 1\% & 87.9 & 74.0 & 67.1 & 23.7 & 19.0 & 17.4 & 44.8 & 37.0 & 25.5 \\
Ours (Voxel-RCNN-based) & 1\% & \textbf{92.5} & \textbf{81.7} & \textbf{77.5} & \textbf{50.7} & \textbf{43.9} & \textbf{42.4} & \textbf{65.2} & \textbf{48.3} & \textbf{42.5} \\ \hline
Voxel-RCNN \cite{voxelrcnn} & 2\% & 89.2 & 76.5 & 71.5 & 44.2 & 40.2 & 34.4 & 56.7 & 39.9 & 37.4 \\
Ours (Voxel-RCNN-based) & 2\% & \textbf{91.6} & \textbf{82.0} & \textbf{77.9} & \textbf{64.9} & \textbf{58.3} & \textbf{50.9} & \textbf{88.0} &\textbf{ 65.7} & \textbf{60.9} \\ \hline
\end{tabular}}
\caption{Experimental results on KITTI dataset based on the Voxel-RCNN detector, where the metrics are the same as \cref{tab:kitti_result}.
}
\label{tab:kitti_result_voxel}
\vspace{-1.4em}
\end{table}

%-------------------------experiments------------------------------------------------
\section{Experiments}
\subsection{Datasets and Evaluation Metrics}
\textbf{KITTI Dataset.} 
Following the state-of-the-art methods \cite{3dioumatch,detmatch}, we evaluate our HSSDA on the KITTI 3D detection benchmark \cite{kitti}, and we use the divided \emph{train} split of 3,712 samples and \emph{val} split of 3,769 samples as a common practice \cite{pvrcnn}. 
Then we sample three different 1\% and 2\% labeled scenes over \emph{train} split based on the released 3DIoUMatch \cite{3dioumatch} splits.
The reported results are averaged over model training on three sampled splits and evaluated on the \emph{val} split.
In addition, the KITTI benchmark has three difficulty levels (easy, moderate, and hard) due to the occlusion and truncation levels of objects.
For fair comparisons, we report the mAP with 40 recall positions, with a 3D IoU threshold of 0.7, 0.5, and 0.5 for the three classes: car, pedestrian, and cyclist, respectively.

\textbf{Waymo Open Dataset.} 
We also evaluate our HSSDA on the Waymo Open Dataset \cite{waymo}, which is one of the biggest autonomous driving datasets, containing 798 sequences (approximately 158k point cloud scenes) for training and 202 sequences (approximately 40k point cloud scenes) for validation, whilst the view of annotations is in full 360$^{\circ}$ field.
We find that even only 1\% of the labeled Waymo scenes contain approximately 3 times as many object annotations as the full KITTI \emph{train} split.
Thus, we sample 1\% of the 798 training sequences (approximately 1.4k point cloud scenes) and report the standard mean average precision (mAP) as well as mAPH, which represent the heading angle factors. 
In addition, the prediction results are split into LEVEL\_1 and LEVEL\_2 for 3D objects including more than five LiDAR points and one LiDAR point, respectively.

\subsection{Implementation Details}
At the training stage, the student network of our HSSDA is end-to-end optimized with the ADAM optimizer and a cosine annealing learning rate\cite{cos}.
As for the weak augmentation for the teacher network, we randomly flip each scene along X-axis and Y-axis with 0.5 probability, and then scale it with a uniformly sampled factor from $[0.91,1.12]$.
Finally, we rotate the point cloud around Z-axis with a random angle sampled from $\left[-\frac{\pi}{4}, \frac{\pi}{4}\right]$.
For the KITTI Dataset, the X-axis and Y-axis are limited in $[0, 70.4]m$ and $[-40, 40]m$ in the shuffle data augmentation, and our HSSDA (PV-RCNN-based) is trained for 80 epochs with the batch size 50.
For the Waymo Open Dataset, the point cloud scene is clipped into $[-75.2, 75.2]m$ for X and Y axes, and training with the batch size 30 for 10 epochs.
We set the value of $\tau_{pair}$ to 0.5 in all experiments according to the evaluation metric in the public datasets.

\subsection{Main Results}
\textbf{KITTI Dataset.}
We first evaluate our proposed model on the popular KITTI dataset.
\cref{tab:kitti_result} lists the results of different methods.
From this table, we can observe that our approach significantly outperforms the state-of-the-art methods.
Specifically, our approach has a remarkable boost in the `Car' class for all settings, which has improvements of 7.4, 5.3, and 4.6 points compared to the PV-RCNN baseline for 1\%, 2\%, and 20\%, respectively. 
Even compared to the recently proposed DetMatch \cite{detmatch} which uses two modalities of LiDAR and RGB, our methods just with LiDAR still have better results for most of the settings.
Besides, we replace the point-voxel-based PV-RCNN 3D detector with a representative voxel-based Voxel-RCNN \cite{voxelrcnn} 3D detector.
The similarly impressive experimental results in \cref{tab:kitti_result_voxel} demonstrate the effectiveness of our HSSDA.

\textbf{Waymo Dataset.}
For the more challenging Waymo Dataset, our approach still has a significant improvement in performance compared to the state-of-the-art methods.
As shown in \cref{tab:waymo_result}, our approach surpasses DetMatch \cite{detmatch}.
It is worth mentioning that the proposed method achieves 29.1 mAP for `Cyclist', which far exceeds the baseline.

%---------------------------ablation study----------------------------------------------
\subsection{Ablation Study}
In this section, we present a series of ablation studies to analyze the effect of our proposed strategies in HSSDA. 
All the experiments are conducted based on the Voxel-RCNN detector with the 2\% KITTI split and evaluated on \emph{val} split due to its fast training speed. 
\cref{tab:abl_main} summarizes the ablation results on our shuffle data augmentation (SDA) and hierarchical supervision (HS) of the teacher network, which provides three levels of supervision: high-confidence level (H\_LEV) pseudo labels, ambiguous level (A\_LEV) pseudo labels, and low-confidence level (L\_LEV) pseudo labels.

\textbf{Effect of the hierarchical supervision.}
It can be found that only considering the high-confidence level pseudo labels will perform better than the baseline as shown in Exp.2 in \cref{tab:abl_main}, but the improvements are limited by the confused background supervision.
The introduction of ambiguous level supervision information can lead to further performance improvements which can be seen in Exp.3.
Furthermore, we can observe that from Exp.4 generating clean scenes via low-confidence supervision can significantly improve the detection accuracy, which indicates the effectiveness of the points removal operation.
Besides, the collaboration of three different levels of supervisions can greatly improve performance, as shown in Exp.5 of \cref{tab:abl_main}.
Those results mean that all hierarchical supervisions have contributions to final performance when they work together. 

\textbf{Effect of the shuffle data augmentation.}
Exp.5 in \cref{tab:abl_main} shows the effect of our shuffle data augmentation strategy. The classes of `Pedestrian' and `Cyclist' have very weak original features due to their small sizes. 
Both of them usually are very hard to detect. 
However, our shuffle data augmentation strategy can significantly improve their performance.
It can be also observed that a slight drop for `Car' may be due to the shuffle data augmentation splitting the original objects, leading to blurred edges for locating.

Our shuffle data augmentation has two hyperparameters: $\mathcal{R}$ and $\mathcal{C}$, which decide the number of scene patches to shuffle.
To evaluate the effect of two hyperparameters, we investigate the performance of the proposed HSSDA with different combinations of $\mathcal{R}$ and $\mathcal{C}$ in \cref{tab:abl_rc}.
We can observe that the model achieves the best result when $\mathcal{R} = \mathcal{C} = 2$ (\emph{i.e.,} the scene grids are split into 4 patches and perform random shuffle).
Hence, we set $\mathcal{R}$ and $\mathcal{C}$ to 2 in all our experiments.

\begin{table}[]
\centering
% \resizebox{7.3cm}{!}{
\resizebox{\columnwidth}{!}{
\begin{tabular}{c|ccc|c|ccc|c}
\hline
\multirow{2}{*}{Exp.} & \multicolumn{3}{c|}{Hierarchical Supervision (HS)} & \multirow{2}{*}{SDA} & \multicolumn{3}{c|}{3D Detection} & \multirow{2}{*}{mAP} \\ \cline{2-4}
 & H\_LEV & A\_LEV & L\_LEV &  & Car & Ped. & Cyc. &  \\ \hline
1 & - & - & - & - & 76.5 & 40.2 & 39.9 & 52.2 \\ \hline
2 & \checkmark & - & - & - & 75.9 & 51.8 & 51.5 & 59.7 \\
3 & \checkmark & \checkmark & - & - & 78.8 & 51.1 & 52.2 & 60.7 \\
4 & \checkmark & \checkmark & \checkmark & - & \textbf{82.4} & 56.0 & 61.3 & 66.5 \\
5 & \checkmark & \checkmark & \checkmark & \checkmark & 82.0 & \textbf{58.3} & \textbf{65.7} & \textbf{68.6} \\ \hline
\end{tabular}
}
\caption{Ablation study of different components in HSSDA. 
}
\label{tab:abl_main}
\end{table}

\begin{table}[]
\centering
\resizebox{5.5cm}{!}{
\begin{tabular}{c|cc|ccc|c}
\hline
\multirow{2}{*}{Method} & \multirow{2}{*}{$\mathcal{R}$} & \multirow{2}{*}{$\mathcal{C}$} & \multicolumn{3}{c|}{3D Detection} & \multirow{2}{*}{mAP} \\ \cline{4-6}
 &  &  & Car & Ped. & Cyc. &  \\ \hline
HS & 1 & 1 & 82.4 & 56.0 & 61.3 & 66.5 \\ \hline
\multirow{4}{*}{HS + SDA} & 1 & 2 & \textbf{82.5} & 57.1 & 64.3 & 67.9 \\
 & 2 & 2 & 82.0 & \textbf{58.3} & 65.7 & \textbf{68.6} \\
 & 2 & 4 & 81.2 & 56.8 & 65.5 & 67.8 \\
 & 4 & 4 & 81.1 & 54.8 & \textbf{65.8} & 67.2 \\ \hline
\end{tabular}
}
\caption{Results of various combinations of $\mathcal{R}$ and $\mathcal{C}$.
}
\label{tab:abl_rc}
\end{table}

\subsection{Quality Analysis}
\begin{figure}[t]
    \centering
    \includegraphics[width=7.5cm]{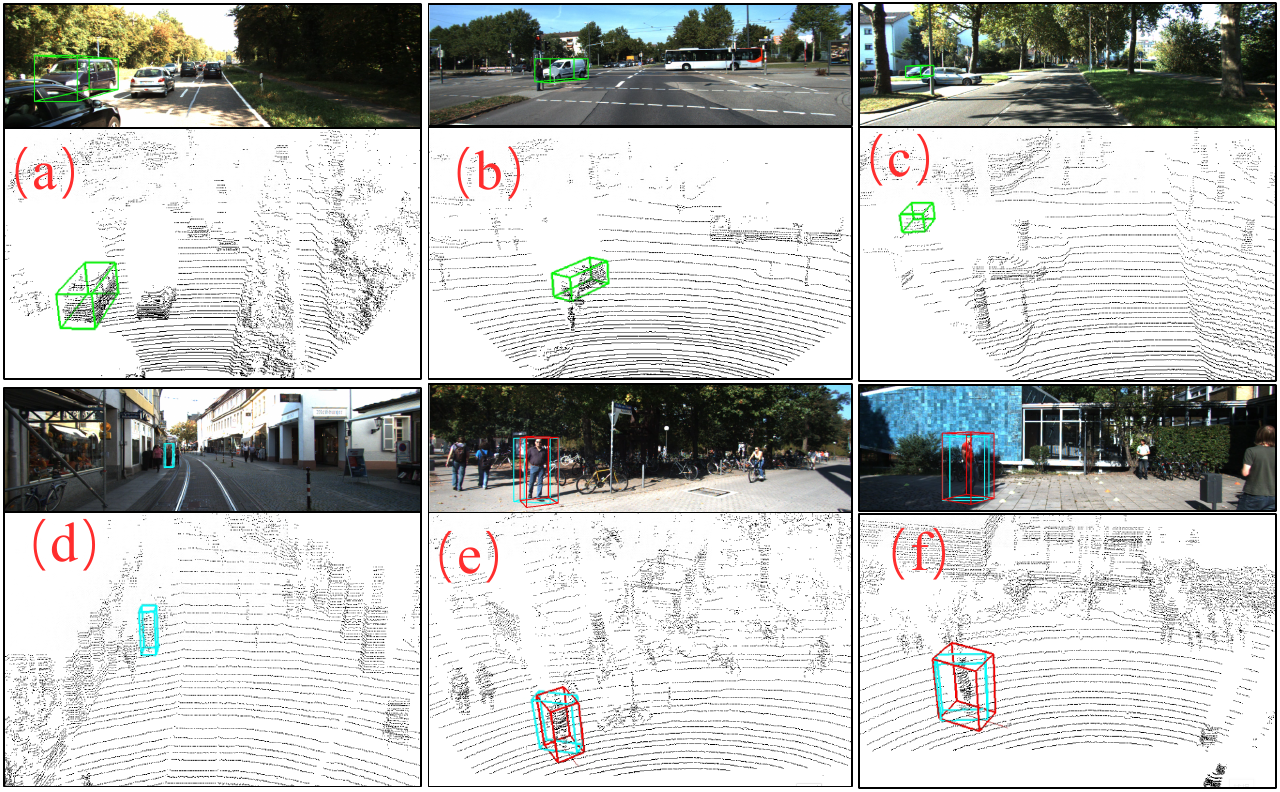}
    \caption{Qualitative analysis of pseudo labels on KITTI.
    For a better view, we only show the objects we are interested in and set the pseudo-labeling car,  pseudo-labeling pedestrian, and ground truth bounding box in green, cyan, and red, respectively, whilst projecting boxes in point cloud back to RGB images. (a) and (b) show the case of category errors, (c) and (d) show the missing-annotated instance of the dataset, and (e) and (f) show the case of poorly localized pseudo labels.
    Best viewed in color.}
    \label{fig:pseudo_label}
    \vspace{-1.4em}
\end{figure}

In this section, we analyze the quality of the generated pseudo labels which play a key role during model training.
First of all, if the 3D IoU between pseudo labels and ground-truth boxes of labeled scenes is bigger than 0.5 with the same class, we regard the pseudo-label as a correctly mined object.
From \cref{tab:abl_pseudo_acc}, we can see the final precision of our high-confidence level pseudo labels for each class on the KITTI dataset is particularly accurate, which indicates the effectiveness of our dual-threshold strategy.
Besides, we provide qualitative results of wrong high-confidence level pseudo labels in \cref{fig:pseudo_label}.
For ease of viewing, we only show the object of one failure case in each scene.
(a) and (b) in \cref{fig:pseudo_label} show that the common failures for `Car' usually occur with similar classes (such as vans and trucks).
Interestingly, our method can reliably mine some real objects which were not annotated in the dataset (see \cref{fig:pseudo_label} (c) and \cref{fig:pseudo_label} (d)).
Additionally, due to the small sizes of `Pedestrian', most of the failure examples are caused by inaccurate localization, as shown in \cref{fig:pseudo_label} (e) and (f) (the ground truth and pseudo-labeling 3D bounding box are drawn in red and cyan.)

\begin{table}[]
\centering
\resizebox{\columnwidth}{!}{
% \resizebox{5.5cm}{!}{
\begin{tabular}{c|ccc}
\hline
\multirow{2}{*}{Category} & \multicolumn{3}{c}{Split setting on KITTI} \\ \cline{2-4} 
 & \multicolumn{1}{c|}{1\%} & \multicolumn{1}{c|}{2\%} & 20\% \\ \hline
Car & \multicolumn{1}{c|}{96.73 ({\color{red}4627}/{\color{blue}4783})} & \multicolumn{1}{c|}{98.69 ({\color{red}4476}/{\color{blue}4535})} & 98.88 ({\color{red}4508}/{\color{blue}4559}) \\ \hline
Pedestrian & \multicolumn{1}{c|}{85.58 ({\color{red}204}/{\color{blue}239})} & \multicolumn{1}{c|}{92.29 ({\color{red}273}/{\color{blue}296})} & 93.67 ({\color{red}74}/{\color{blue}79}) \\ \hline
Cyclist & \multicolumn{1}{c|}{95.53 ({\color{red}107}/{\color{blue}112})} & \multicolumn{1}{c|}{95.00 ({\color{red}114}/{\color{blue}120})} & 96.33 ({\color{red}105}/{\color{blue}109}) \\ \hline
\end{tabular}
}
\caption{Final precision of high-confidence level pseudo labels on `Car', `Pedestrian' and `Cyclist' classes with different SSL settings. The blue and red numbers represent the total  and correctly number of mined pseudo labels, respectively.
}
\label{tab:abl_pseudo_acc}
\end{table}

%------------------------------------------------------------------------
\section{Conclusion}
In this paper, we propose a novel teacher-student-based method for 3D semi-supervised object detection, called HSSDA.
Through the dual-threshold strategy in the teacher network, we can provide hierarchical supervision to effectively train the student network, while eliminating the negative impact of missing-mined objects of unlabeled scenes.
In addition, the shuffle data augmentation strategy shuffles the input and unshuffles the feature blocks to strengthen the feature representation ability of the student network.
Extensive experiments validate the superiority of our method in challenging datasets.
Our HSSDA can train any 3D detector which consists of a backbone and detection head.

\noindent\textbf{Limitations.} 
Our HSSDA designs a dynamic dual-threshold strategy that determines the optimal threshold in a global view.
So, we need to use the teacher network for additional validation, which requires more training time.
In addition, the shuffle data augmentation may split the complete objects resulting in blurred edges and locating.

\noindent\textbf{Acknowledgments.} 
This work is supported in part by the National Key R\&D Program of China (2022YFA1004100), and in part by the National Natural Science Foundation of China (No.62176035, 62201111, 62036007), the Science and Technology Research Program of Chongqing Municipal Education Commission under Grant (No.KJZD-K202100606), the Chongqing graduate Research Innovation Project (CYB22249), the CQUPT Ph.D. Innovative Talents Project (BYJS202105), and the Macao Science and Technology Development Fund (061/2020/A2).

%%%%%%%%% REFERENCES
{\small
\bibliographystyle{ieee_fullname}
\bibliography{egbib}
}

\newpage
\clearpage
\appendix

\begin{table}[htp]
\centering
\resizebox{\columnwidth}{!}{
% \resizebox{0.5\textwidth}{!}{
\begin{tabular}{c|c|c|c|c|c|c}
\hline
\begin{tabular}[c]{@{}c@{}}1\% Data \\ ($\sim$ 1.4k scenes)\end{tabular} & \begin{tabular}[c]{@{}c@{}}Veh. \\ (LEVEL 1)\end{tabular} & \begin{tabular}[c]{@{}c@{}}Veh. \\ (LEVEL 2)\end{tabular} & \begin{tabular}[c]{@{}c@{}}Ped. \\ (LEVEL 1)\end{tabular} & \begin{tabular}[c]{@{}c@{}}Ped. \\ (LEVEL 2)\end{tabular} & \begin{tabular}[c]{@{}c@{}}Cyc. \\ (LEVEL 1)\end{tabular} & \begin{tabular}[c]{@{}c@{}}Cyc. \\ (LEVEL 2)\end{tabular} \\ \hline
Voxel-RCNN \cite{voxelrcnn} & 49.02/48.03 & 42.36/41.50 & 41.16/32.81 & 34.73/27.66 & 5.84/5.61 & 5.62/5.40 \\
Ours (Voxel-RCNN-based) & \textbf{54.89}/\textbf{54.06} & \textbf{48.28}/\textbf{47.53} & \textbf{43.86}/\textbf{37.84} & \textbf{36.59}/\textbf{31.56} & \textbf{17.47}/\textbf{16.73} & \textbf{16.72}/\textbf{16.01} \\ \hline
\end{tabular}}
\caption{
Results on the Waymo for the Voxel-RCNN detector.
}
\label{tab:re_waymo}
\end{table}

\begin{table}[htp]
\centering
% \resizebox{0.5\textwidth}{!}{
\resizebox{\columnwidth}{!}{
\begin{tabular}{c|c|ccc|ccc|ccc}
\hline
\multirow{2}{*}{Model} & \multirow{2}{*}{Data} & \multicolumn{3}{c|}{3D Detection (Car)} & \multicolumn{3}{c|}{3D Detection (Ped.)} & \multicolumn{3}{c}{3D Detection (Cyc.)} \\ \cline{3-11} 
 &  & Easy & Mod & Hard & Easy & Mod & Hard & Easy & Mod & Hard \\ \hline
PV-RCNN \cite{pvrcnn} & 100\% & \textbf{92.10} & 84.36 & \textbf{82.48} & \textbf{63.12} & 54.84 & \textbf{51.78} & 89.10 & 70.38 & 66.01 \\ \hline
PV-RCNN \cite{pvrcnn} with SDA & 100\% & 91.91 & \textbf{84.57} & 82.31 & 62.83 & \textbf{55.49} & 51.04 & \textbf{89.68} & \textbf{71.09} & \textbf{66.71} \\ \hline 
\end{tabular}}
\caption{Ablation study of SDA in the fully supervised framework.}
\label{tab:re_sda}
\end{table}

\section{Additional Experimental Results}
\noindent\textbf{(1) Additional experiments on the Waymo Dataset.}
We additionally test the Voxel-RCNN~\cite{voxelrcnn} on 1\% of the Waymo~\cite{waymo} dataset, and the results in \cref{tab:re_waymo} still show the superiority of our method, which validates its generalization.

\noindent\textbf{(2) If the shuffle data augmentation (SDA) strategy is also effective for full supervision training ?}
To verify the effect of the SDA on fully-supervised 3D object detector, we inset the SDA into the PV-RCNN \cite{pvrcnn} and the results are listed in \cref{tab:re_sda}, which shows that the superiority of SDA in the supervised framework is not as obvious as in the semi-supervised framework.
This is due to that the design of the strong augmentation in the student branch module has two main purposes: (1) strong enough to make a significant difference with weakly augmented samples of the teacher branch and (2) not too strong to ensure effective supervision information transmission.

\section{Visualization of Dynamic Dual-Threshold}
\begin{figure}[htp]
    \centering
    \includegraphics[width=7.7cm]{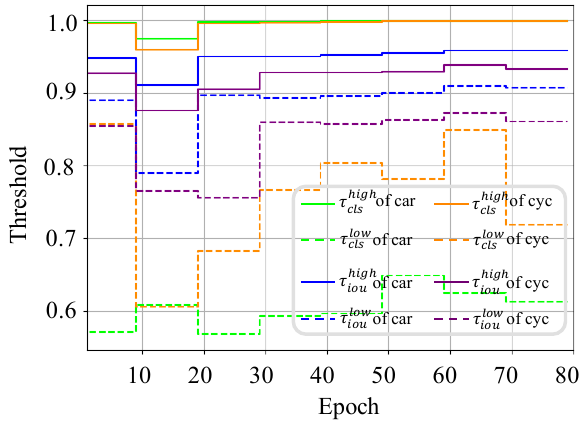}
    \caption{Visualization curve of the dynamic dual-threshold during training.
    }
    \label{fig:re_threshold}
\end{figure}

To better understand the dual-threshold hierarchical supervision in intuitive, we visualize the dynamic threshold changes during the training process in \cref{fig:re_threshold}, where a solid line of a certain color represents a high threshold, and the dotted line of the same color represents a low threshold.

\section{Discuss on Other Augmentation Methods for Point Cloud}
The Mixup-based \cite{mixup} augmentation methods have been extensively studied in the field of image classification and widely applied in 2D semi-supervised object detection \cite{stac} task. Following this idea, there have been several explorations in point cloud tasks as well. PointMixup~\cite{pointmixup} first applied the idea of Mixup to point cloud and achieved linear interpolation through the optimal allocation. 
Mix3D~\cite{mix3d} balances global contextual information and local geometric information to achieve high-performance models. 
In addition, PointCutMix~\cite{pointcutmix} proposes two different ways of replacing points to mix two point clouds. The latest SageMix explores salient regions in two point clouds and smoothly combines them into a continuous shape.
However, these methods mainly focus on point cloud classification and segmentation tasks.
For outdoor 3D object detection task, objects are usually naturally separated~\cite{pointrcnn}, and merging two point cloud scenes will cause overlaps between objects (\eg, two vehicles are rarely overlapped in 3D reality). 
Therefore, to the best of our knowledge, the above Mixup-based point cloud augmentation methods cannot be directly applied to detection tasks, which is the direction for our future research.

\end{document}